\pdfoutput=1

\documentclass[11pt]{article}

\usepackage{EMNLP2023}

\usepackage{times}
\usepackage{latexsym}
\usepackage{lscape}
\usepackage{graphicx}
\usepackage{url}
\usepackage{amsmath}
\usepackage{booktabs}
\usepackage{multirow}
\usepackage{nicefrac}
\usepackage[nameinlink,capitalize,noabbrev]{cleveref}
\usepackage{amssymb}
\usepackage{todonotes}
\usepackage{xcolor}

\usepackage[T1]{fontenc}

\usepackage[utf8]{inputenc}

\usepackage{microtype}

\usepackage{subcaption}
\usepackage{inconsolata}
\usepackage{bm}

\definecolor{Red}{rgb}{1.0,0.0,0.3}
\definecolor{Green}{rgb}{0,0.6,0}
\definecolor{Blue}{rgb}{0,0.3,1.0}
\definecolor{Grey}{rgb}{0.6,0.6,0.6}

\title{Not all layers are equally as important: Every Layer Counts BERT}

\author{Lucas Georges Gabriel Charpentier \and David Samuel \\
  University of Oslo, Language Technology Group \\
  {\tt \{lgcharpe, davisamu\}@ifi.uio.no} \\
}

\begin{document}
\maketitle

\begin{abstract}
This paper introduces a novel modification of the transformer architecture, tailored for the data-efficient pretraining of language models. This aspect is evaluated by participating in the BabyLM challenge, where our solution won both the \textsc{strict} and \textsc{strict-small} tracks. Our approach allows each transformer layer to select which outputs of previous layers to process. The empirical results verify the potential of this simple modification and show that not all layers are equally as important.

\end{abstract}

\section{Introduction}


Modern language models (LLMs), with their deep architectures and large parameter counts, have displayed outstanding performance on a wide range of tasks. Their ability to understand, generate, and manipulate human language has been groundbreaking \citep{devlin-etal-2019-bert, 2020t5, NEURIPS2020_1457c0d6}. However, this success largely relies on \textit{vast amounts of unsupervised data} that these models need for pretraining, requiring extensive computational power and time. While this is feasible for high-resource languages like English, it becomes a bottleneck for languages with limited data resources \citep{joshi-etal-2020-state}. Moreover, the environmental and economic costs of such massive training regimens are growing concerns \citep{strubell-etal-2019-energy, thompson2020computational}. 

The BabyLM challenge tries to address these concerns by providing a shared experimental ground for efficient language modelling \citep{warstadt-et-al-2023-babylm}. All models submitted to this shared task have to be trained on a restricted text corpus of 10M and 100M words -- in the \textsc{strict-small} and \textsc{strict} tracks, respectively. The challenge pushes the boundaries of what is possible with data-efficient language model pretraining.

\renewcommand{\arraystretch}{1.2}
\begin{table}[t!]
\resizebox{\columnwidth}{!}{%
\begin{tabular}{@{}l@{\hspace{-2em}}rrr@{\hspace{3em}}r@{}}
{\small\textsc{strict-small} track (10M words)}\\
\toprule
\textbf{Model} & \textbf{BLiMP} & \textbf{GLUE} & \textbf{MSGS} & \textbf{Average}\\\midrule
ELC-BERT \textit{(ours)} & \textbf{75.8}	& \textbf{73.7} & \textbf{29.4} &	\textbf{65.9} \\
MLSM & 72.4	& 70.6 & 17.2 &	60.8 \\
Contextualizer & 74.3 & 69.6 & 12.7 & 60.5 \\
Baby Llama & 69.8 & 67.6 & 24.7 & 60.1 \\
Too Much Information & 75.7 & 70.9 & 3.9 & 59.9 \\
\bottomrule
\\
{\small\textsc{strict} track (100M words)}\\
\toprule
\textbf{Model} & \textbf{BLiMP} & \textbf{GLUE} & \textbf{MSGS} & \textbf{Average}\\\midrule
ELC-BERT \textit{(ours)} & \textbf{82.8}	& 78.3 & 47.2 &	\textbf{74.3} \\
Contextualizer & 79.0	& 72.9 & \textbf{58.0} &	73.0 \\
BootBERT & 82.2 & \textbf{78.5} & 27.7 & 70.2 \\
MSLM & 76.2 & 73.5 & 21.4 & 64.4 \\
Bad babies & 77.0 & 67.2 & 23.4 & 63.4 \\
\bottomrule
\end{tabular}%
}
\caption{\label{tab:dynabench}
The DynaBench scores of the BabyLM challenge \citep{warstadt-et-al-2023-babylm}, the table shows the top 5 submissions in the \textsc{strict-small} and \textsc{strict} tracks. Higher scores are better, the best results in each evaluation suite are boldfaced.
}
\end{table}

In response to this challenge, we present a novel modification to the well-established transformer architecture \citep{vaswani2017attention}.  Instead of traditional residual connections, our model allows each layer to \textit{selectively} process outputs from the preceding layers. This flexibility leads to intriguing findings: not every layer is of equal significance to the following layers. Thus, we call it the `Every Layer Counts' BERT (ELC-BERT).

The BabyLM challenge provided us with a robust benchmark to evaluate the efficacy of ELC-BERT. Our approach emerged as the winning submission in both the \textsc{strict} and \textsc{strict-small} tracks (\cref{tab:dynabench}), which highlights the potential of layer weighting for future low-resource language modelling.

\begin{figure*}[!t]
    \includegraphics[width=\textwidth]{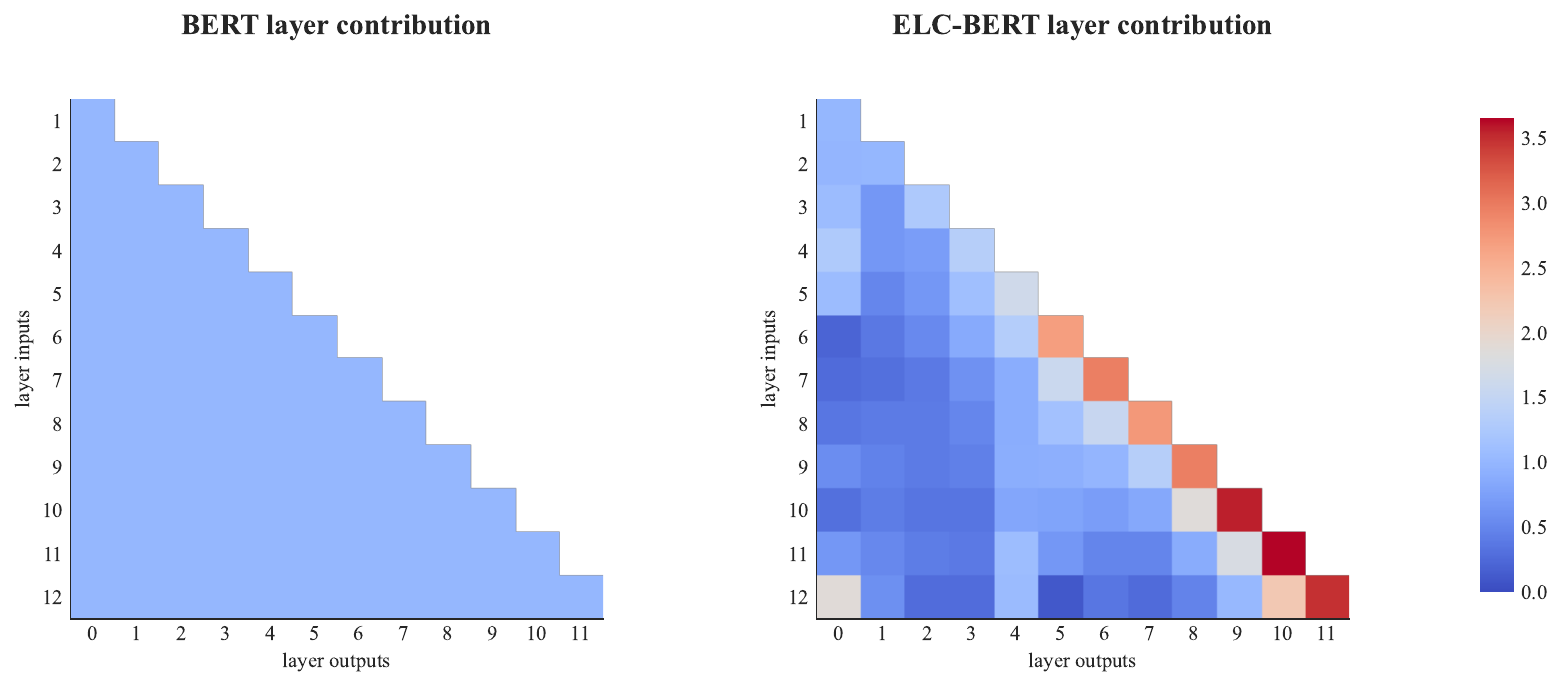}
    \caption{Every layer can select which outputs from previous layers it wants as its input, these heatmaps show the weights given to each previous layer output. The unit weights of the BERT model (and of any standard transformer-based model) are inferred from \cref{eq:residual}. The right heatmap shows the $\alpha$ weights of the normalized ELC-BERT variant; for clear visual comparison between the two models, we rescale the $\alpha$ weights so that the $k$th row sums to $k$. Note that the layer 0 is the embedding layer, as in \cref{eq:embedding}.}
    \label{fig:layer-weights}
\end{figure*}

Transparent and open-source language modelling is necessary for safe future development of this field. We release the full source code, together with the pre-trained ELC-BERT models, online.\footnote{\url{https://github.com/ltgoslo/elc-bert}}





\section{Related work}

\paragraph{Residual and highway networks.} While the predecessor of residual models, highway networks, used a conditional gating mechanism to weigh layers \citep{NIPS2015_215a71a1}, modern residual networks (including transformers) simply weigh all layers equally \citep{he2016residual, vaswani2017attention}. Our work reintroduces layer weights into residual models -- but without the computational cost of a gating mechanism.

\paragraph{Layer importance.} The difference between various layers inside pre-trained language models has been extensively studied \citep{jawahar-etal-2019-bert, tenney-etal-2019-bert, niu-etal-2022-bert}. Different layers process different linguistic phenomena, thus their \textit{importance} for downstream tasks varies -- this has been successfully utilized by learning layer weights during finetuning, for example in ULMFiT \citep{howard-ruder-2018-universal} or UDify \citep{kondratyuk-straka-2019-75}. Following this direction, our system uses layer weights in the finetuning as well as in the pretraining phase.

\paragraph{ReZero transformer.} A related approach to ours was proposed by \newcite{bachlechner2021rezero}. In that paper, the authors experimented with scaling the output of each layer. They showed that by initializing the scaling parameter to zero, their `ReZero transformer' model tends towards setting the scale to $\nicefrac{1}{N}$ (where $N$ is the number of layers). Our approach can be considered as a generalization of this method -- in ELC-BERT, every layer weights the outputs of previous layers \textit{individually}.




\section{ELC-BERT layer weighting} \label{sec:methods}

We modify the residual connections inside the transformer architecture so that every layer can select which outputs from previous layers it wants to process -- instead of always taking a simple sum of all preceding layers, as done in the Transformer \citep{vaswani2017attention} and in most works that use a variant of this architecture. This modification allows the model to form a complex inter-layer structure, as visible from \cref{fig:layer-weights}.

\paragraph{Transformer definition.} To be more specific, we first formally define a \textit{transformer encoder} as a function that maps subword indices $\bm{x}$ onto subword probabilities $\bm{y}$. First, $\bm{x}$ is embedded into a vector representation $\bm{h}^0_{\text{out}}$, which is then processed by $N$ layers consisting of attention and multi-layer-perceptron (MLP) modules. Finally, $\bm{y}$ is produced by processing the final hidden representation with a language-modelling head. Formally for $n \in \{1,\,\dots N\}$:
\begin{align}
    \bm{h}^0_{\text{out}} &\gets \operatorname{embedding}(\bm{x}), \label{eq:embedding}\\
    \bm{h}^n_{\text{out}} &\gets \operatorname{att}(\bm{h}^n_{\text{in}}) + \operatorname{mlp}\!\left({\bm{h}^n_{\text{in}} + \operatorname{att}(\bm{h}^n_{\text{in}})}\right), \label{eq:transformer-layer}\\
    \bm{y} &\gets \operatorname{LM\_head}(\bm{h}^{N}_{\text{out}}). \label{eq:lm-head}
\end{align}

\paragraph{The original residual connection.} The original transformer definition by \newcite{vaswani2017attention} can be recovered by simply assigning
\begin{align}
    \bm{h}^n_{\text{in}} \gets \bm{h}^{n-1}_{\text{out}} + \bm{h}^{n-1}_{\text{in}}.
    \label{eq:residual}
\end{align}
\noindent This recurrent assignment can also be rewritten as $\bm{h}^n_{\text{in}} \gets \sum^{n-1}_{i=0}{\bm{h}^i_{\text{out}}}$, which highlights the implicit assumption of residual models that the output from every previous layer is equally important. 

\paragraph{Layer weighting.} In our formulation, we make two changes to the original definition: {\color{Red}(i)} the residual connections in all MLP modules are removed, {\color{Blue}(ii)} the input to every layer is a convex combination of outputs from previous layers. Specifically, we replace \cref{eq:transformer-layer} and \cref{eq:residual} by:
\begin{align}
    \bm{h}^n_{\text{out}} &\gets \operatorname{att}(\bm{h}^n_{\text{in}}) + \operatorname{mlp}\!\left({\color{Red}\operatorname{att}(\bm{h}^n_{\text{in}})}\right), \label{eq:new-transformer-layer}\\
    \bm{h}^n_{\text{in}} &\gets \sum^{n-1}_{i=0}{{\color{Blue}\alpha_{i,n} }\bm{h}^i_{\text{out}}}, \label{eq:new-residual}
\end{align}

\noindent where $\sum^{n-1}_{i=0}{\alpha_{i,n} = 1}$. This constraint is satisfied by a softmax transformation of the raw learnable layer weights $\hat{\alpha}_{*,n} \in \mathbb{R}^n$ into $\alpha_{*,n}$. $\hat{\alpha}_{*,n}$ is initialized as a zero vector except for the value of $\hat{\alpha}_{n-1,n}$ set to one, to bias the weight towards the input from the previous layer. 

\section{Training}


\paragraph{LTG-BERT backbone.} We base our models around LTG-BERT \citep{samuel-etal-2023-trained}. This model has been specifically optimized for pretraining on small text corpora, similar to the one provided by BabyLM. We adopt all of their architectural modifications, their language modelling objective as well as all other pretraining settings. We also use the raw LTG-BERT (without our layer weighting) as a strong baseline in the following evaluation. Details on the pretraining hyperparameters can be found in \cref{tab:hyperparams}.

\paragraph{BabyLM pretraining corpus.}

We pretrain all language models on a corpus from the BabyLM challenge \citep{warstadt-et-al-2023-babylm}. The goal of this challenge is to shed more light on data-efficient language modelling and on the question of human language acquisition. Thus, the organizers have constructed a small-scale text corpus of the same type and quantity that children learn from.

Specifically, the shared task consists of three tracks: \textsc{strict}, \textsc{strict-small} and \textsc{loose}. We participate in the first two tracks, where the submissions have to be pre-trained only on the BabyLM corpus, which corpus contains about 100M words in the \textsc{strict} track and about 10M words in the \textsc{strict-small} track. We adopt the preprocessing pipeline from \newcite{samuel2023bootbert} for unifying the format of texts from this corpus.

\begin{table}
\resizebox{\columnwidth}{!}{
\begin{tabular}{@{}l@{\hspace{-4em}}rrrr@{}}
{\small\textsc{strict-small} track (10M words)}\\
\toprule
\textbf{Model} & \textbf{BLiMP} & \textbf{Supp.} & \textbf{MSGS} & \textbf{GLUE}\\
\midrule
OPT\textsubscript{125m} & 62.6  & 54.7 & -0.64$^{\pm 0.1}$ & 68.3$^{\pm3.3}$ \\ 
RoBERTa\textsubscript{base} & 69.5 & 47.5 & -0.67$^{\pm 0.1}$ & 72.2$^{\pm1.9}$ \\ 
T5\textsubscript{base}  & 58.8 & 43.9 & -0.68$^{\pm 0.1}$ & 64.7$^{\pm1.3}$ \\
LTG-BERT\textsubscript{small} & \textbf{80.6} & \textbf{69.8} & -\textbf{0.43}$^{\pm 0.4}$ & 74.5$^{\pm 1.5}$ \\ 
ELC-BERT\textsubscript{small} & 80.5 & 67.9 & -0.45$^{\pm 0.2}$ & \textbf{75.3$^{\pm 2.1}$} \\ 
\bottomrule
\\
{\small\textsc{strict} track (100M words)}\\
\toprule
\textbf{Model} & \textbf{BLiMP} & \textbf{Supp.} & \textbf{MSGS} & \textbf{GLUE}\\
\midrule
OPT\textsubscript{125m} & 75.3  & 67.8 & -0.44$^{\pm 0.1}$ & 73.0$^{\pm3.9}$ \\ 
RoBERTa\textsubscript{base} & 75.1 & 42.4 & -0.66$^{\pm 0.3}$ & 74.3$^{\pm0.6}$ \\ 
T5\textsubscript{base}  & 56.0 & 48.0 & -0.57$^{\pm 0.1}$ & 75.3$^{\pm1.1}$ \\
LTG-BERT\textsubscript{base} & \textbf{85.8} & \textbf{76.8} & -0.42$^{\pm 0.2}$ & 77.9$^{\pm 1.1}$ \\ 
ELC-BERT\textsubscript{base} & 85.3 & 76.6 & \textbf{-0.26$^{\pm 0.5}$} & \textbf{78.3$^{\pm 3.2}$} \\ 
\bottomrule
\end{tabular}%
}
\caption{\label{tab:res_baby}
Results for the BabyLM challenge suite of evaluation datasets -- BLiMP, supplemental dataset to BLiMP, MSGS and (Super)GLUE. We compare the results of our submitted model (ELC-BERT\textsubscript{biased}) to the backbone model (LTG-BERT\textsubscript{base}) and the baselines given by the organizers of the challenge on the \textsc{strict} dataset. On the \textsc{strict-small} dataset, we compare a variation (ELC-BERT\textsubscript{zero}) of small size to the backbone model and baselines.
}
\end{table}

\section{Results}

This section provides the results of the empirical evaluation of ELC-BERT. First, we compare our method to baselines, then we perform an ablation study of different ELC-BERT variations, and finally, we take a deeper look into the learnt layer weights.

\subsection{BabyLM challenge evaluation}


We adopt the BabyLM evaluation pipeline for all comparisons.\footnote{\url{https://github.com/babylm/evaluation-pipeline}} The pipeline itself is an adaptation of \newcite{eval-harness} and it aims to provide a robust evaluation of syntactic and general language understanding. 

The syntactic understanding is measured by the Benchmark of Linguistic Minimal Pairs \citep[BLiMP \& BLiMP supplemental;][]{warstadt2020blimp} and the Mixed Signals Generalization Set 
\citep[MSGS;][]{warstadt2020learning}. The general natural language understanding is measured by GLUE and SuperGLUE \citep{wang-etal-2018-glue, NEURIPS2019_4496bf24}. All of these benchmarks use filtered subsets of the original datasets (provided by the organizers), which means that they are not directly comparable to previous literature. If applicable, we divide the training set into a train-development split and report the mean/std statistics over multiple runs on the former validation split.


\paragraph{BLiMP.}
This benchmark tests zero-shot preference of grammatical sentences. From the \textsc{strict} results in \cref{tab:res_baby}, we see that ELC-BERT outperforms the baseline models by a fair margin on this task. 
However, if we look at the LTG-BERT baseline, we see that our model slightly underperforms it (by 0.5 percentage points). \cref{blimp} provides a more in-depth comparison of the models.

If we now look at the supplemental scores, we see a very similar trend to the BLiMP results: our model outperforms the baseline RoBERTa model by 24.4 p.p. while slightly underperforming against the LTG-BERT model by 0.2 p.p. \cref{suppl} shows a breakdown of the aggregated scores. 



\begin{figure}
    \centering
    \includegraphics[width=\columnwidth]{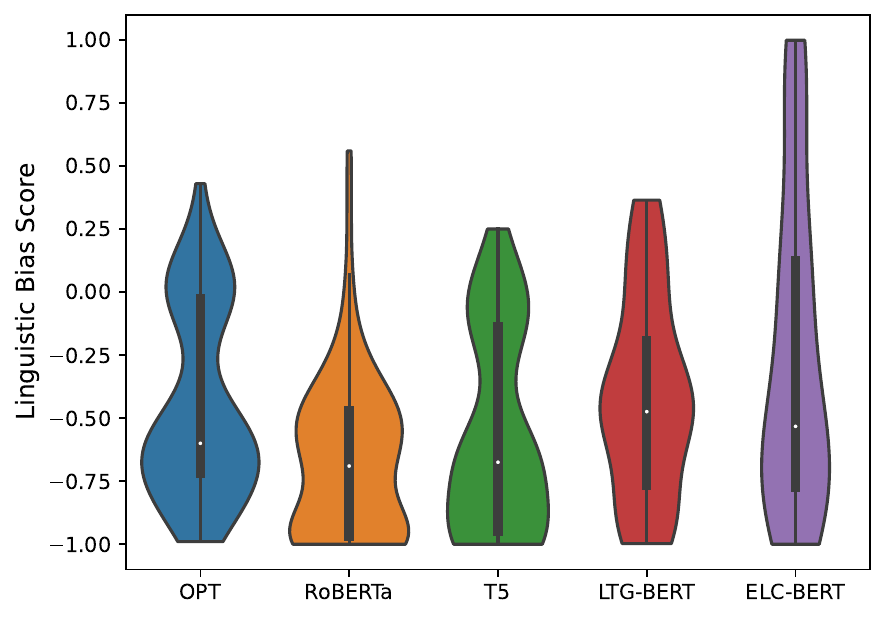}
    \caption{Violin plots of each model's Linguistic Bias Scores (LBS) and the base model. The white dot shows the median LBS and the edge of the boxes are the 1\textsuperscript{st} and 3\textsuperscript{rd} quartiles. The width of the violins shows the density of results at that score.}
    \label{fig:msgs_gen_baseline}
\end{figure}

\paragraph{GLUE.} 

A standard LM benchmark that tests the ability to be finetuned for general language understanding tasks. Focusing on the results in \cref{tab:res_baby}, we see that our model outperforms both the encoder baseline 
and the LTG-BERT model 
in the \textsc{strict} and \textsc{stric-small} tracks. The improvement against LTG-BERT is rather modest and could be caused by random variation. If we look at \cref{glue} we see that the variation is greatly affected by the WSC task -- 
ignoring it, we get a score of $80.49^{\pm{1.44}}$ for our model and $79.52^{\pm{1.13}}$ for LTG-BERT. 


\paragraph{MSGS.} 

Finally, this benchmark evaluates the preference towards linguistic explanations over spurious surface explanations. 
For the aggregated \textsc{strict} MSGS results of \cref{tab:res_baby}, the comparison appears unclear due to the large standard deviation. However, a closer inspection reveals that ELC-BERT \textit{significantly} outperforms LTG-BERT by 0.16 LBS points.\footnote{Using the Almost Stochastic Order (ASO) significance test from \citet{dror-etal-2019-deep} and \citet{del2018optimal} (calculated using \citet{ulmer2022deep}), we get a $\varepsilon_{\min}$ of 0.2 at a confidence level of 0.95 which implies that there is a high likelihood that ELC-BERT is better than LTG-BERT.} \cref{fig:msgs_gen_baseline} and \cref{tab:msgs} shows a detailed view on the score distribution. 



\paragraph{Shared task results.}

The official Dynabench results for the top-5 models for the \textsc{strict} and \textsc{strict-small} track can be found in \cref{tab:dynabench}. Looking first at the \textsc{strict} track results, we see that our model achieves the highest total score and BLiMP score, while we are second for GLUE and MSGS. On the \textsc{strict-small} track our model performs best on all benchmarks and by a substantial margin for all benchmarks. 



\subsection{Model variations}


We compare the following modifications of the ELC-BERT architecture from \cref{sec:methods}:
\begin{enumerate}
    \item \textbf{Zero initialization}: The layer weights are all initialized as zeros, without any bias towards the previous layer. This model also uses the residual MLP input from \cref{eq:transformer-layer}. This variation is used in the \textsc{strict-small} track. 
    \item \textbf{Strict normalization}: This follows the previous variant with every $\bm{h}^i_{\text{out}}$ normalized to a unit vector. 
    \item \textbf{Weighted output}: Follows the first variant and the input to the LM head is a weighted sum of all layers. To be more concrete, we replace \cref{eq:lm-head} by $\bm{y} \gets \operatorname{LM\_head}\!\left(\sum^{N}_{i=0}{\alpha_{i,N+1} \bm{h}^i_{\text{out}}}\right)$. 
\end{enumerate}

\begin{table}
\resizebox{\columnwidth}{!}{
\begin{tabular}{@{}lrrrr@{}}
\toprule
\textbf{Model} & \textbf{BLiMP} & \textbf{Supp.} & \textbf{MSGS} & \textbf{GLUE}\\
\midrule
ELC-BERT & 85.3 & 76.6 & -0.26$^{\pm 0.5}$ & 78.3$^{\pm 3.2}$ \\ 
\,\,\, + zero initialization & 84.9 & \textbf{78.5} & -0.38$^{\pm 0.3}$ & \textbf{79.4$^{\pm 1.0}$} \\ 
\,\,\, + normalization & 85.1 & 76.0 & \textbf{-0.13$^{\pm 0.4}$} & 78.2$^{\pm 3.3}$ \\ 
\,\,\, + weighted output & \textbf{86.1} & 76.0 & -0.28$^{\pm 0.2}$ & 78.2$^{\pm 0.6}$ \\ 
\bottomrule
\end{tabular}%
}
\caption{\label{tab:res_elc}
Results for the BabyLM challenge suite of evaluation datasets. We compare the performance of different variants of our model to the one submitted to the BabyLM challenge as well as the backbone model LTG-BERT on the \textsc{strict} dataset.
}
\end{table}

\paragraph{Evaluation.}

Based on \cref{tab:res_elc}, we see that different variations have varying effects on the evaluation scores. 

\begin{figure}
    \centering
    \includegraphics[width=\columnwidth]{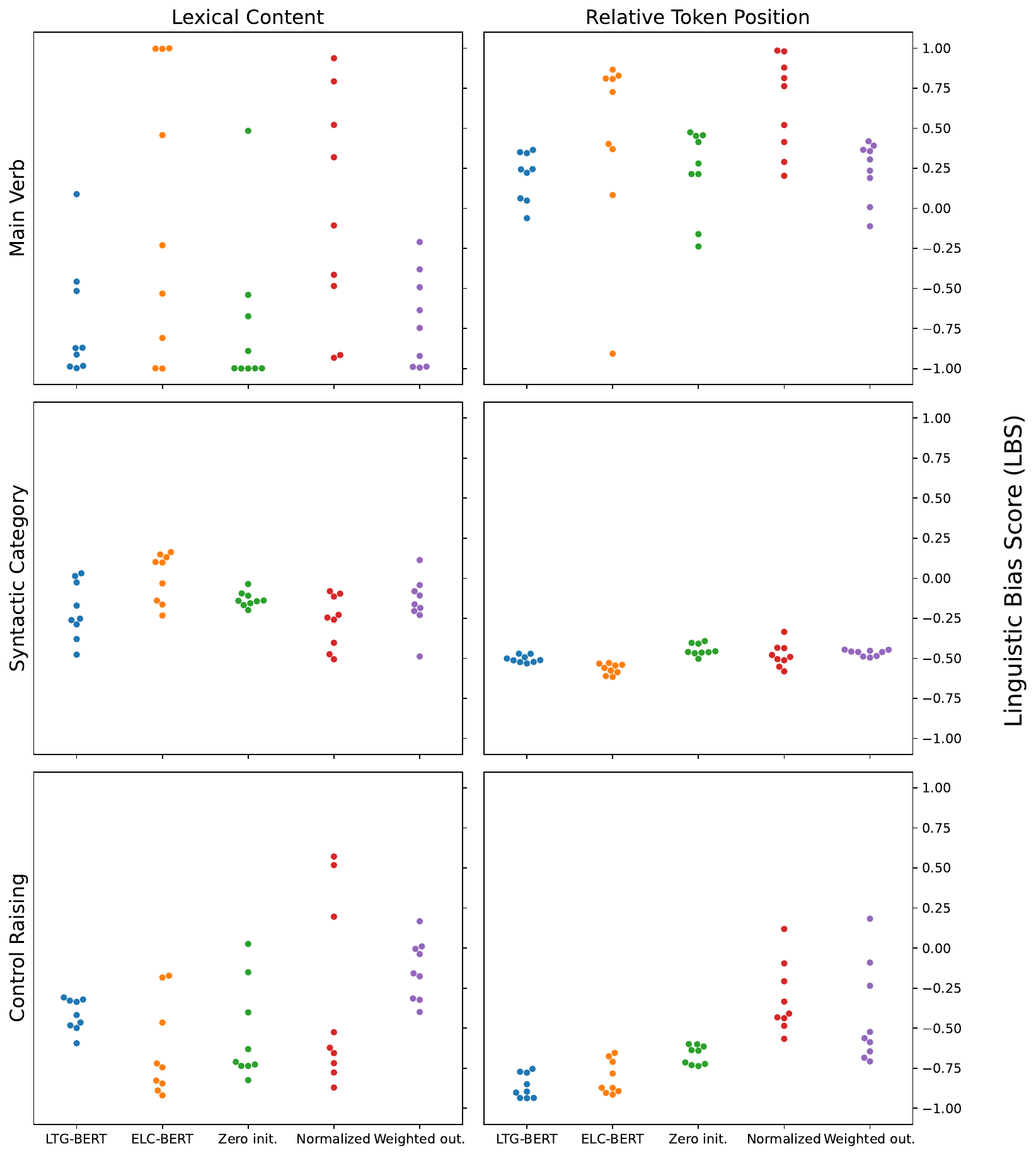}
    \caption{Detailed LBS for each model and each combination of surface and linguistic features. The Y-axis (Main Verb, Syntactic Category, and Control Raising) show the linguistic features, while the X-axis (Lexical Content, Relative Token Position) represent the surface features. Each dot represents a different fine-tuned model.}
    \label{fig:msgs}
\end{figure}

When changing the $\hat{\alpha}$ initialization to zero, we see a significant increase in performance on both the BLiMP Supplemental and the GLUE benchmarks.
\footnote{The increase in performance on the GLUE benchmark is significant when using the ASO significance test both against the original ELC-BERT and the backbone model LTG-BERT. Against both models, we get a $\varepsilon_{\min}$ of 0, indicating a very strong likelihood that the zero variation is better than ELC-BERT and LTG-BERT on GLUE} However, the model suffers in performance on both the BLiMP and MSGS.\footnote{This is a significant decrease with an $\varepsilon_{\min}$ of 0.28 that ELC-BERT is better.} Overall, we see that this variation leads to better zero-shot and fine-tuning results while biasing the model more towards spurious surface features rather than linguistic features, as can be seen in \cref{fig:msgs}. 

If we then focus on the normalization variation, we see that it underperforms in all benchmarks but one, MSGS, where it significantly performs better by 0.13 LBS points,\footnote{Significant with an $\varepsilon_{\min}$ of 0.31.} 
as can be seen in more detail in \cref{fig:msgs}.

Finally, when looking at our weighted output variation, we see a substantial gain in performance on the BLiMP benchmark 
while the results on MSGS and GLUE are similar
, and the results on Supplemental BLiMP slightly decrease. 
More detailed results on all these benchmarks can be found in \cref{app:detail}.



\subsection{Layer importance}

The empirical evaluation suggests that learnable layer weights are a simple but effective architectural change -- but how do these learnt weights look like? In this section, we investigate the $\alpha$ values of the normalized ELC-BERT variant.\footnote{The interpretation of $\alpha$ weights in a non-normalized variant is difficult due to different magnitudes of layer outputs.} 



Looking at the importance matrix of ELC-BERT in \cref{fig:layer-weights}, we posit that the first 5 layers focus on surface-level information found in the embedding layer explaining its enhanced importance for the embedding layer. The next 5 layers (6-10) focus on more linguistic features by virtually ignoring the first 4 layers (0-3) and focusing primarily on the previous three layers as well as layers 4 and 5 to get some transformed information from the embedding layer. Layer 11 does much the same but focuses more on Layer 4, potentially trying to obtain some surface knowledge found in it. Finally, Layer 12 behaves similarly to Layer 11 but also puts high importance (3\textsuperscript{rd} most) on the embedding layer. This is most likely to recuperate some surface information lost in previous layers to pass to the language modelling head.



\section{Conclusion}


In this paper, we proposed a novel and simple modification of the transformer architecture for language modelling. We empirically tested the efficacy of our approach by participating in the BabyLM challenge -- a shared task for data-efficient language modelling. Our submission ranked first on both tracks that we participated in. A more detailed evaluation shows that, when compared to a strong baseline, our approach reliably performs better on (Super)GLUE tasks. The evaluation on MSGS suggests that our approach is more likely to prefer linguistic features over spurious surface features, and the BLiMP benchmarks show comparable performance to the baseline. Finally, our proposed modification shows that the assumption that all layers are equally important is incorrect, and a more complex layer structure helps the model.

\section*{Acknowledgements}
The efforts described in the current paper were funded by the HPLT project (High-Performance Language Technologies; coordinated by Charles University). The computations were performed on resources provided through Sigma2 -- the national research infrastructure provider for High-Performance Computing and large-scale data storage in Norway.

\bibliography{anthology,custom}
\bibliographystyle{acl_natbib}

\clearpage
\onecolumn
\appendix

\section{Pre-training details} \label{app:pre}

\begin{table*}[ht!]
\centering
\small
\begin{tabular}{@{}lccc@{}}
\toprule
\textbf{Hyperparameter} & \textbf{Base} & \textbf{Small} & \textbf{Small (Submitted Model)} \\ \midrule
Number of parameters    & 98M & 24M & 24M \\
Number of layers        & 12 & 12 & 12        \\
Hidden size             & 768 & 384 & 384         \\
FF intermediate size    & 2\,048 & 1\,024 & 1\,024  \\
Vocabulary size         & 16\,384 & 6\,144 & 6\,144         \\
Attention heads         & 12 & 6 & 6          \\
Hidden dropout                 & 0.1   & 0.1 & 0.1       \\
Attention dropout       & 0.1   & 0.1 & 0.1         \\
Training steps          & 15\,625 & 15\,625 & 31\,250    \\
Batch size              & 32\,768 & 32\,768 & 8\,096     \\
Initial Sequence length         & 128 & 128 & 128       \\
Initial Sequence length         & 512 & 512 & 512       \\
Warmup ratio            & 1.6\% & 1.6\% & 1.6\%        \\
Initial learning rate   & 0.01 & 0.0141 & 0.005         \\
Final learning rate     & 0.001 & 0.00141 & 0.005         \\
Learning rate scheduler & cosine & cosine & cosine        \\
Weight decay            & 0.1 & 0.4 & 0.4 \\
Layer norm $\epsilon$   & 1e-7 & 1e-7 & 1e-7         \\
Optimizer               & LAMB & LAMB & LAMB        \\
LAMB $\epsilon$         & 1e-6 & 1e-6 & 1e-6        \\
LAMB $\beta_1$          & 0.9  & 0.9 & 0.9        \\
LAMB $\beta_2$          & 0.98   & 0.98  & 0.98     \\
Gradient clipping       & 2.0  & 2.0 & 2.0         \\ \bottomrule
\end{tabular} %
\caption{Pre-training hyperparameters for the small-sized models (trained on \textsc{strict-small}) and for the base-sized models (trained on the \textsc{strict} track).}
\label{tab:hyperparams}
\end{table*}

\newpage

\section{Fine-tuning details} \label{app:fine}

For the fine-tuning experiments, we will run multiple seeds and (for MSGS) multiple learning rates, to be able to get a more robust comparison of model performance. The detailed hyperparameters for fine-tuning can be found in \cref{tab:glue-hyperparams}.

\subsubsection{GLUE}

To finetune, we will use 5 different seeds: 12, 642, 369, 1267, and 2395. We will use a validation set to find our best model with early-stopping, and then test our model on a test set (here the validation set is 10\% of the training sets from \url{https://github.com/babylm/evaluation-pipeline} and the test set is their validation set).

\subsubsection{MSGS}

To finetune, we use three different random seeds: 12, 369, and 2395, as well as three different learning rates: 1e-5, 2e-5, and 3e-5. In addition, we train for 5 epochs, with a batch size of 16 with no early stopping.

\begin{table*}[ht!]
\small
\centering
\begin{tabular}{@{}lccc@{}}
\toprule
\multirow{2}{*}{\textbf{Hyperparameter}} & \textbf{QQP, MNLI} & \textbf{CoLA, RTE, WSC} & \multirow{2}{*}{\textbf{MSGS}} \\
    & \textbf{QNLI, SST-2} & \textbf{MRPC, MultiRC} & \\\midrule

Batch size             & 32 & 16 & 16  \\
Number of epochs       & 10 & 10 & 5   \\
Dropout                & 0.1& 0.1 & 0.1 \\
Warmup steps           & 10\%  & 1\%                     & 6\% \\
Peak learning rate    & 5e-5     & 7e-5                     & \{1e-5, 2e-5, 3e-5\}      \\
Learning rate decay  & cosine    & cosine                   & linear  \\
Weight decay           & 0.1   & 0.1                     & 0.1 \\
Optimizer             & AdamW        & AdamW                    & AdamW \\
Adam $\epsilon$     & 1e-8    & 1e-8                     & 1e-8 \\
Adam $\beta_1$       & 0.9        & 0.9                      & 0.9    \\
Adam $\beta_2$     & 0.999        & 0.999                    & 0.999  \\ \bottomrule
\end{tabular}
\caption{Hyperparameters for fine-tuning the GLUE, SuperGLUE task and MSGS tasks. We use the same hyperparameters for all ELC-BERT models, not performing any per-model hyperparameter search. The values for MSGS are adopted from \citep{warstadt2020learning}. For all models, we measure the statistics over 5 random seeds for GLUE tasks: 12, 642, 369, 1267, and 2395; and 3 seeds for MSGS tasks: 12, 369, and 2395}
\label{tab:glue-hyperparams}
\end{table*}

\newpage

\section{BabyLM dataset}

\cref{tab:data} is a detailed overview of the BabyLM dataset:

\begin{table*}[ht!]
    \centering
    \resizebox{\textwidth}{!}{%
    \begin{tabular}{@{}llrrr@{}}
    \toprule
    & & \multicolumn{2}{c}{\textbf{\# Words}} & \\\cmidrule(lr){3-4}
    Dataset & Domain & \textsc{strict-small} & \textsc{strict} & Proportion \\
    \midrule
    CHILDES \citep{macwhinney2000childes} & Child-directed speech & 0.44M & 4.21M & 5\% \\
    British National Corpus (BNC),\textsuperscript{1} dialogue portion & Dialogue & 0.86M & 8.16M & 8\% \\
    Children's Book Test \citep{hill-2016-cbt} & Children's books & 0.57M & 5.55M & 6\% \\
    Children's Stories Text Corpus\textsuperscript{2} & Children's books & 0.34M & 3.22M & 3\% \\
    Standardized Project Gutenberg Corpus \citep{gerlach-2018-gutenberg} & Written English & 0.99M & 9.46M & 10\% \\
    OpenSubtitles \citep{lison-tiedemann-2016-opensubtitles2016} & Movie subtitles & 3.09M & 31.28M & 31\% \\
    QCRI Educational Domain Corpus (QED; \citealp{abdelali-etal-2014-qed}) & Educational video subtitles & 1.04M & 10.24M & 11\% \\
    Wikipedia\textsuperscript{3} & Wikipedia (English) & 0.99M & 10.08M & 10\% \\
    Simple Wikipedia\textsuperscript{4} & Wikipedia (Simple English) & 1.52M & 14.66M & 15\% \\
    Switchboard Dialog Act Corpus \citep{Stolcke-etal:2000} & Dialogue & 0.12M & 1.18M & 1\% \\
    \midrule
    \emph{Total} & -- & 9.96M & 98.04M & 100\% \\
    \bottomrule
    \end{tabular}}
    \caption{The contents of datasets for the the \textsc{strict} and \textsc{strict-small} tracks; the table is taken from \newcite{warstadt-et-al-2023-babylm}. \textsuperscript{1}\footnotesize\url{http://www.natcorp.ox.ac.uk}\ \ \ \textsuperscript{2}\url{https://www.kaggle.com/datasets/edenbd/children-stories-text-corpus}\ \ \ \textsuperscript{3}\url{https://dumps.wikimedia.org/enwiki/20221220/}\ \ \ \textsuperscript{4}\url{https://dumps.wikimedia.org/simplewiki/20221201/}}
    \label{tab:data}
\end{table*}

\section{Detailed Results} \label{app:detail}

This section breaks down the aggregate scores of the benchmarks into their composing tasks. It also describes or name each task

\subsection{BLiMP}

The BabyLM challenge uses the BLiMP benchmark \citep{warstadt2020blimp} to evaluate the syntactic understanding of the models. Our detailed results can be found in \cref{blimp}. Its composing tasks are as follows (with descriptions taken from \citet{warstadt2020blimp}):
\begin{itemize}
    \item \textsc{Anaphor Agreement} (AA): the requirement that reflexive pronouns like \textit{herself} (also known as anaphora) agree with their antecedents in person, number, gender, and animacy.
    \item \textsc{Argument structure} (AS): the ability of different verbs to appear with different types of arguments. For instance, different verbs can appear with a direct object, participate in the causative alternation, or take an inanimate argument.
    \item \textsc{Binding} (B): the structural relationship between a pronoun and its antecedent.
    \item \textsc{Control/raising} (CR): syntactic and semantic differences between various types of predicates that embed an infinitival VP. This includes control, raising, and \textit{tough}-movement predicates.
    \item \textsc{Determiner-noun agreement} (DNA): number agreement between demonstrative determiners (e.g., \textit{this/these}) and the associated noun.
    \item \textsc{Ellipsis} (E): the possibility of omitting expressions from a sentence. Because this is difficult to illustrate with sentences of equal length, our paradigms cover only special cases of noun phrase ellipsis that meet this constraint.
    \item \textsc{Filler-gap} (FG): dependencies arising from phrasal movement in, for example, \textit{wh}-questions.
    \item \textsc{Irregular forms} (IF): irregular morphology on English past participles (e.g., \textit{awoken}).
    \item \textsc{Island effects} (IE): restrictions on syntactic environments where the gap in a filler-gap dependency may occur.
    \item \textsc{NPI licensing} (NL): restrictions on the distribution of \textit{negative polarity items} like \textit{any} and \textit{ever} limited to, for example, the scope of negation and \textit{only}.
    \item \textsc{Quantifiers} (Q): restrictions on the distribution of quantifiers.  Two such restrictions are covered: superlative quantifiers (e.g., \textit{at least}) cannot be embedded under negation, and definite quantifiers and determiners cannot be subjects in existential-\textit{there} constructions.
    \item \textsc{Subject-verb agreement} (SVA): subjects and present tense verbs must agree in number.
\end{itemize}

\begin{table*}[ht!]
\centering
\scriptsize
\begin{tabular}{lccccccccccccc}
\toprule
\textbf{Model} & \textbf{AA} & \textbf{AS} & \textbf{B} & \textbf{CR} & \textbf{DNA} & \textbf{E} & \textbf{FG} & \textbf{IF} & \textbf{IE} & \textbf{NL} & \textbf{Q} & \textbf{SVA} & \textbf{Average}\\
\midrule
\multicolumn{5}{@{}l}{\tiny{\textsc{strict} (100M words)}}    \\
OPT\textsubscript{125M} & 94.9& 73.8& 73.8& 72.2& 93.1& 80.5& 73.6& 80.8& 57.8& 51.6& 74.5& 77.3& 75.3\\
RoBERTa\textsubscript{base} & 89.5& 71.3& 71.0& 67.1& 93.1& 83.8& 68.0& 89.6& 54.5& 66.3& 70.3& 76.2& 75.1\\
T5\textsubscript{base} & 66.7& 61.2& 59.4& 59.8& 53.8& 49.1& 70.0& 75.5& 43.6& 45.6& 34.2& 53.2& 56.0\\
LTG-BERT\textsubscript{base} & \textbf{96.1} & 79.5 & \textbf{77.1} & 80.3 & 95.4 & \textbf{91.7} & 87.8 & 94.5 & 79.8 & 84.4 & 72.2 & \textbf{91.2} & 85.8 \\[1em]
ELC-BERT\textsubscript{base} & 92.8 & \textbf{81.2} & 74.0 & 79.2 & \textbf{96.0} & \textbf{91.7} & 87.1 & 93.6 & \textbf{83.9} & 83.5 & 70.2 & 90.8 & 85.3 \\
\,\,\, + zero initialization & 93.8 & 79.1 & 73.6 & 79.8 & 95.5 & 91.0 & 87.1 & 93.3 & 78.8 & 84.8 & \textbf{73.5} & 88.7 & 84.9 \\
\,\,\, + normalization & 93.0 & 79.1 & 74.6 & 79.8 & 95.6 & \textbf{91.7} & 87.4 & 93.9 & 82.0 & 83.7 & 71.3 & 89.1 & 85.1 \\
\,\,\, + weighted output & 94.7 & 80.7 & 75.7 & \textbf{81.3} & 95.7 & 91.6 & \textbf{88.9} & \textbf{95.9} & 83.2 & \textbf{85.7} & 69.2 & 91.1 & \textbf{86.1} \\
\midrule
\multicolumn{5}{@{}l}{\tiny{\textsc{strict-small} (10M words)}}    \\
OPT\textsubscript{125M} & 63.8 & 70.6 & 67.1 & 66.5 & 78.5 & 62.0& 63.8 & 67.5 & 48.6 & 46.7 & 59.6 & 56.9& 62.6\\
RoBERTa\textsubscript{base} & 81.5& 67.1& 67.3& 67.9& 90.8& 76.4& 63.5& 87.4& 39.9& 55.9& 70.5& 65.4& 69.5\\
T5\textsubscript{base} & 68.9& 63.8& 60.4& 60.9& 72.2& 34.4& 48.2& 77.6& 45.6& 47.8& 61.2& 65.0& 58.8\\[1em]
ELC-BERT\textsubscript{small} & 89.5 & \textbf{72.5} & \textbf{68.1} & \textbf{72.6} & 93.4 & 87.4 & \textbf{80.6} & \textbf{91.0} & \textbf{67.9} & 79.4 & \textbf{75.2} & \textbf{88.7} & \textbf{80.5} \\
\bottomrule
\end{tabular}
\caption{\label{blimp}
BLiMP results for models trained both on the 100M (above the mid-horizontal line) and the 10M (below the mid-horizontal line) Baby LM dataset. The \textbf{bold} results represent the best model for the task. The metric used to measure is accuracy. The results are in percentage.
}
\end{table*}

\subsection{BLiMP Supplemental}

\begin{table*}[ht!]
\centering
\scriptsize
\begin{tabular}{lccccccccccccc}
\toprule
\textbf{Model} & \textbf{Hypernym} & \textbf{QA Congruence Easy} & \textbf{QA Congruence Tricky} & \textbf{Subject Aux Inversion} & \textbf{Turn Talking} & \textbf{Average}\\
\midrule
\multicolumn{5}{@{}l}{\tiny{\textsc{strict} (100M words)}}    \\
OPT\textsubscript{125M} & 46.3 & 76.5 & 47.9 & 85.3 & 82.9 & 67.8 \\
RoBERTa\textsubscript{base} & 50.8 & 34.4 & 34.5 & 45.6 & 46.8 & 42.4 \\
T5\textsubscript{base} & \textbf{51.1} & 45.3 & 25.5 & 69.2 & 48.9 & 48.0 \\
LTG-BERT\textsubscript{base} & 47.0 & 90.6 & 60.6 & 90.7 & 92.1 & 76.8 \\[1em]
ELC-BERT\textsubscript{base} & 47.3 & 85.9 & 63.0 & 94.5 & 92.1 & 76.6 \\
\,\,\, + zero initialization & 47.1 & \textbf{92.2} & \textbf{64.2} & 95.9 & \textbf{93.2} & \textbf{78.5} \\
\,\,\, + normalization & 46.1 & 85.9 & 59.4 & \textbf{96.5} & 92.1 & 76.0 \\
\,\,\, + weighted output & 48.6 & 87.5 & 57.6 & 96.2 & 90.4 & 76.0 \\
\midrule
\multicolumn{5}{@{}l}{\tiny{\textsc{strict-small} (10M words)}}    \\
OPT\textsubscript{125M} & \textbf{50.0} & 54.7 & 31.5 & 80.3 & 57.1 & 54.7 \\
RoBERTa\textsubscript{base} & 49.4 & 31.3 & 32.1 & 71.7 & 53.2 & 47.5 \\
T5\textsubscript{base} & 48.0 & 40.6 & 21.2 & 64.9 & 45.0 & 43.9 \\[1em]
ELC-BERT\textsubscript{small} & 48.0 & \textbf{73.4} & \textbf{43.6} & \textbf{90.0} & \textbf{84.3} & \textbf{67.9} \\
\bottomrule
\end{tabular}
\caption{\label{suppl}
BLiMP supplemental results for models trained both on the 100M (above the mid-horizontal line) and the 10M (below the mid-horizontal line) Baby LM dataset. The \textbf{bold} results represent the best model for the task. The metric used to measure is accuracy. The results are in percentage.
}
\end{table*}

\subsection{GLUE}

The BabyLM challenge involves slightly modified GLUE and SuperGLUE benchmarks. It uses only a subset of the subtasks, the datasets are filtered so that they do not contain out-of-vocabulary words, and it sometimes uses non-standard metrics. Our detailed results can be found in \cref{glue}. We list all subtasks and their metrics below:

\begin{itemize}\itemsep0em 
    \item \textbf{Boolean Questions} \citep[BoolQ;][]{clark-etal-2019-boolq}, a yes/no Q/A dataset evaluated with accuracy.
    \item \textbf{Corpus of Linguistic Acceptability} \citep[CoLA;][]{warstadt-etal-2019-neural} evaluated with accuracy (originally evaluated with the Matthews correlation coefficient \citep[MCC;][]{MATTHEWS1975442}).
    \item \textbf{The Multi-Genre Natural Language Inference Corpus} \citep[MNLI;][]{williams-etal-2018-broad}. Its development set consists of two parts: \textit{matched}, sampled from the same data source as the training set, and \textit{mismatched}, which is sampled from a different domain. Both parts are evaluated with accuracy.
    \item \textbf{The Microsoft Research Paraphrase Corpus} \citep[MRPC;][]{dolan-brockett-2005-automatically}, evaluated with both F\textsubscript{1}-score (originally also evaluated with accuracy).
    \item \textbf{Multi-Sentence Reading Comprehension} \citep[MultiRC;][]{khashabi-etal-2018-looking}, a multiple choice question answering dataset, evaluated with accuracy (originally evaluated with the exact match accuracy (EM) and F\textsubscript{1}-score (over all answer options)).
    \item \textbf{Question-answering Natural Language Inference} (QNLI) constructed from the Stanford Question Answering Dataset \citep[SQuAD;][]{rajpurkar-etal-2016-squad}, evaluated with accuracy.
    \item \textbf{The Quora Question Pairs} (QQP),\footnote{\url{https://quoradata.quora.com/First-Quora-Dataset-Release-Question-Pairs}} evaluated with F\textsubscript{1}-score (originally evaluated with accuracy).
    \item \textbf{The Stanford Sentiment Treebank} \citep[SST-2;][]{socher-etal-2013-recursive}, evaluated with accuracy.
    \item \textbf{The Recognizing Textual Entailment datasets} \citep[RTE;][]{10.1007/11736790_9, rte2, giampiccolo-etal-2007-third, Bentivogli09thefifth}, evaluated with accuracy.
    \item \textbf{Winograd Schema Challenge} \citep[WSC;][]{10.5555/3031843.3031909} evaluated with accuracy.
\end{itemize}

\begin{table}[ht!]
\resizebox{\textwidth}{!}{
\begin{tabular}{@{}lcccccccccccc@{}}
\toprule
\textbf{Model} & \textbf{CoLA} & \textbf{SST-2} & \textbf{MRPC} & \textbf{QQP} & \textbf{MNLI\textsubscript{m}} & \textbf{MNLI\textsubscript{mm}} & \textbf{QNLI} & \textbf{RTE} & \textbf{BoolQ} & \textbf{MultiRC} & \textbf{WSC} & \textbf{Average}\\

\midrule
\multicolumn{5}{@{}l}{\small{\textsc{strict} (100M words)}}    \\[1em]

OPT\textsubscript{125m}       & 74.9$^{\pm0.6}$ & 87.7$^{\pm0.7}$ & 81.9$^{\pm0.7}$ & 84.3$^{\pm0.1}$ & 75.7$^{\pm0.3}$ & 77.0$^{\pm0.3}$ & 82.8$^{\pm0.8}$ & 58.6$^{\pm2.9}$ & 66.4$^{\pm0.7}$ & 61.5$^{\pm0.8}$ & 52.3$^{\pm12.5}$ & 73.0$^{\pm3.9}$ \\

RoBERTa\textsubscript{base}   & 75.6$^{\pm0.3}$ & 88.3$^{\pm0.6}$ & 84.0$^{\pm0.5}$ & 85.5$^{\pm0.2}$ & 77.4$^{\pm0.4}$ & 78.3$^{\pm0.3}$ & 83.6$^{\pm0.2}$ & 50.7$^{\pm1.5}$ & 67.7$^{\pm0.7}$ & 64.3$^{\pm0.5}$ & 61.4$^{\pm0.0}$ & 74.3$^{\pm0.6}$ \\

T5\textsubscript{base}        & 76.7$^{\pm0.9}$ & 89.0$^{\pm0.8}$ & 85.2$^{\pm1.1}$ & 86.2$^{\pm0.1}$ & 77.9$^{\pm0.3}$ & 78.7$^{\pm0.3}$ & 84.7$^{\pm0.9}$ & 55.4$^{\pm2.2}$ & 67.7$^{\pm1.5}$ & 65.7$^{\pm0.8}$ & 61.0$^{\pm1.1}$ & 75.3$^{\pm1.1}$ \\

LTG-BERT\textsubscript{base} & 82.7$^{\pm 0.8}$ & 92.0$^{\pm 0.4}$ & 87.4$^{\pm 0.7}$ & 87.9$^{\pm 0.1}$ & 83.0$^{\pm 0.4}$ & 83.4$^{\pm 0.5}$ & 89.1$^{\pm 0.5}$ & 54.7$^{\pm 2.4}$ & 68.4$^{\pm 0.5}$ & 66.0$^{\pm 1.4}$ & 61.4$^{\pm 0.0}$ & 77.9$^{\pm 1.1}$ \\[1em]

ELC-BERT\textsubscript{base} & 82.6$^{\pm 0.5}$ & 91.9$^{\pm 1.1}$ & \textbf{89.3$^{\pm 0.6}$} & 88.0$^{\pm 0.1}$ & 83.6$^{\pm 0.1}$ & 83.3$^{\pm 0.2}$ & 89.4$^{\pm 0.4}$ & 60.0$^{\pm 2.8}$ & 70.5$^{\pm 1.5}$ & \textbf{66.2$^{\pm 2.2}$} & 56.4$^{\pm 9.4}$ & 78.3$^{\pm 3.2}$ \\

\,\,\, + zero initialization & 82.0$^{\pm 0.7}$ & \textbf{92.4$^{\pm 0.4}$} & 88.8$^{\pm 1.5}$ & \textbf{88.2$^{\pm 0.1}$} & \textbf{84.4$^{\pm 0.3}$} & \textbf{84.5$^{\pm 0.3}$} & \textbf{90.5$^{\pm 0.5}$} & \textbf{63.0$^{\pm 1.5}$} & \textbf{72.6$^{\pm 1.0}$} & 65.8$^{\pm 1.1}$ & 61.4$^{\pm 0.0}$ & \textbf{79.4$^{\pm 1.0}$} \\

\,\,\, + normalization & \textbf{83.1$^{\pm 0.4}$} & 91.9$^{\pm 0.4}$ & 88.6$^{\pm 1.3}$ & 88.0$^{\pm 0.1}$ & 84.1$^{\pm 0.2}$ & 84.3$^{\pm 0.2}$ & \textbf{90.5$^{\pm 0.4}$} & 56.2$^{\pm 2.4}$ & 72.0$^{\pm 1.5}$ & 64.9$^{\pm 0.6}$ & 56.9$^{\pm 10.2}$ & 78.2$^{\pm 3.3}$ \\

\,\,\, + weighted output & 82.6$^{\pm 0.6}$ & 91.7$^{\pm 1.2}$ & 87.8$^{\pm 1.2}$ & 87.9$^{\pm 0.1}$ & 84.0$^{\pm 0.4}$ & 84.0$^{\pm 0.3}$ & 89.4$^{\pm 0.3}$ & 55.2$^{\pm 5.5}$ & 71.0$^{\pm 0.8}$ & 64.4$^{\pm 0.8}$ & \textbf{61.7$^{\pm 0.5}$} & 78.2$^{\pm 0.6}$ \\

\midrule
\multicolumn{5}{@{}l}{\small{\textsc{strict-small} (10M words)}}    \\[1em]

OPT\textsubscript{125m}       & 69.0$^{\pm0.5}$ & 85.4$^{\pm0.9}$ & 80.0$^{\pm1.8}$ & 80.3$^{\pm0.3}$ & 69.5$^{\pm0.2}$ & 71.0$^{\pm0.5}$ & 71.5$^{\pm0.7}$ & 51.3$^{\pm2.1}$ & 66.2$^{\pm1.5}$ & 56.5$^{\pm2.0}$ & 50.8$^{\pm10.3}$ & 68.3$^{\pm3.3}$ \\

RoBERTa\textsubscript{base}   & 70.4$^{\pm0.4}$ & 85.6$^{\pm0.3}$ & 82.2$^{\pm0.4}$ & 83.5$^{\pm0.2}$ & 72.5$^{\pm0.4}$ & 74.4$^{\pm0.3}$ & 80.3$^{\pm0.7}$ & \textbf{56.8}$^{\pm5.5}$ & 65.8$^{\pm2.9}$ & 61.2$^{\pm1.5}$ & \textbf{61.7}$^{\pm0.5}$ & 72.2$^{\pm1.9}$ \\

T5\textsubscript{base}        & 76.7$^{\pm0.9}$ & 69.4$^{\pm0.1}$ & 81.4$^{\pm0.6}$ & 76.8$^{\pm0.3}$ & 57.3$^{\pm0.8}$ & 58.6$^{\pm1.1}$ & 64.3$^{\pm0.9}$ & 52.7$^{\pm2.4}$ & 63.4$^{\pm1.6}$ & 48.4$^{\pm1.4}$ & 60.0$^{\pm2.2}$ & 64.7$^{\pm1.3}$ \\

LTG-BERT\textsubscript{small} & \textbf{77.6}$^{\pm 0.8}$ & 88.8$^{\pm 0.8}$ & 82.3$^{\pm 0.4}$ & 85.8$^{\pm 0.2}$ & 78.0$^{\pm 0.2}$ & 78.8$^{\pm 0.4}$ & 85.0$^{\pm 0.2}$ & 53.7$^{\pm 4.1}$ & 64.8$^{\pm 2.1}$ & \textbf{64.1}$^{\pm 0.3}$ & 60.5$^{\pm 1.0}$ & 74.5$^{\pm 1.5}$ \\[1em]

ELC-BERT\textsubscript{small} & 76.1$^{\pm 1.0}$ & \textbf{89.3$^{\pm 0.5}$} & \textbf{85.0$^{\pm 1.8}$} & \textbf{86.7$^{\pm 0.3}$} & \textbf{79.2$^{\pm 0.3}$} & \textbf{79.9$^{\pm 0.2}$} & \textbf{85.8$^{\pm 0.4}$} & 55.4$^{\pm 2.6}$ & \textbf{69.3$^{\pm 2.0}$} & 62.2$^{\pm 1.0}$ & 59.0$^{\pm 5.4}$ & \textbf{75.3$^{\pm 2.1}$} \\


\bottomrule
\end{tabular}
}
\caption{\label{glue}
A subset of GLUE results (defined by the Baby LM challenge) for both the models trained on 100M and 10M words. All the results indicate the model accuracy for the task except for MRPC and QQP where the results are based on the F1-score of the positive class. To obtain the standard deviation, each model is trained with 5 seeds, and the average accuracy/F1-score is reported. The results are reported in percentage. The \textbf{bold} result indicates the best model for each dataset.
}
\end{table}

\subsection{MSGS}

The BabyLM challenge uses a reduced set of the MSGS benchmark \citep{warstadt2020learning} to evaluate whether the model biases linguistic features or surface features. A score of 1 means only using the linguistic features, while a score of -1 is surface features only. \cref{tab:msgs} shows the detailed results of the reduced MSGS benchmark. The first 5 results (MVC to RTPC) are controls, checking whether the model can recognize the feature, while the next six evaluate whether the model biases linguistic or surface features. To evaluate the performance we use the Mathews Correlation Coefficient (MCC), also called Linguistic Bias Score (LBS) for the last six tasks. The surface features in this dataset are (definitions taken from \citet{warstadt2020learning}):
\begin{itemize}
    \item \textsc{Lexical content} (LC): This feature is 1 \textit{iff} the sentence contains \textit{the}.
    \item \textsc{Relative token position} (RTP): This feature is 1 when \textit{the} precedes \textit{a}, and 0 when \textit{a} precedes \textit{the}.
\end{itemize}
The linguistic features are (definitions taken from \citet{warstadt2020learning}):
\begin{itemize}
    \item \textsc{Main verb} (MV): This feature is 1 \textit{iff} the sentence's main verb is in the \textit{-ing} form.
    \item \textsc{Control/raising} (CR): This feature has value 1 \textit{iff} the sentence contains the control construction.
    \item \textsc{Syntactic category} (SC): This feature is 1 \textit{iff} the sentence contains an adjective.
\end{itemize}

\begin{table}[ht!]
\resizebox{\textwidth}{!}{
\begin{tabular}{@{}lccccccccccc@{}}
\toprule
\textbf{Model} & \textbf{MVC} & \textbf{CRC} & \textbf{SCC} & \textbf{LCC} & \textbf{RTPC} & \textbf{MVLC} & \textbf{MVRTP} & \textbf{CRLC} & \textbf{CRRTP} & \textbf{SCLC} & \textbf{SCRTP}\\
\midrule
\multicolumn{5}{@{}l}{\small{\textsc{strict} (10M words)}}    \\[1em]

OPT\textsubscript{125M} & \textbf{1.00$^{\pm 0.00}$} & 0.88$^{\pm 0.04}$ & 0.36$^{\pm 0.06}$ & 0.14$^{\pm 0.04}$ & 0.83$^{\pm 0.03}$ & -0.55$^{\pm 0.12}$ & -0.88$^{\pm 0.06}$ & \textbf{-0.02}$^{\pm 0.08}$ & -0.73$^{\pm 0.05}$ & \textbf{0.11}$^{\pm 0.13}$ & -0.59$^{\pm 0.04}$ \\

RoBERTa\textsubscript{base} & \textbf{1.00$^{\pm 0.00}$} & 0.75$^{\pm 0.12}$ & 0.57$^{\pm 0.22}$ & \textbf{1.00$^{\pm 0.00}$} & 0.92$^{\pm 0.07}$ & -0.87$^{\pm 0.41}$ & -0.89$^{\pm 0.13}$ & -0.37$^{\pm 0.34}$ & -0.54$^{\pm 0.13}$ & -0.70$^{\pm 0.27}$ & -0.61$^{\pm 0.19}$ \\

T5\textsubscript{base} & \textbf{1.00$^{\pm 0.00}$} & 0.82$^{\pm 0.05}$ & 0.56$^{\pm 0.05}$ & \textbf{1.00$^{\pm 0.00}$} & 0.90$^{\pm 0.05}$ & -1.00$^{\pm 0.00}$ & -0.95$^{\pm 0.03}$ & -0.13$^{\pm 0.10}$ & -0.61$^{\pm 0.03}$ & 0.03$^{\pm 0.12}$ & -0.73$^{\pm 0.04}$ \\

LTG-BERT\textsubscript{base} & \textbf{1.00$^{\pm 0.00}$} & 0.83$^{\pm 0.07}$ & 0.65$^{\pm 0.08}$ & \textbf{1.00$^{\pm 0.00}$} & 0.50$^{\pm 0.06}$ & -0.72$^{\pm 0.36}$ & 0.20$^{\pm 0.15}$ & -0.42$^{\pm 0.10}$ & -0.86$^{\pm 0.08}$ & -0.20$^{\pm 0.18}$ & -0.50$^{\pm 0.02}$ \\[1em]

ELC-BERT\textsubscript{base} & \textbf{1.00$^{\pm 0.00}$} & 0.89$^{\pm 0.10}$ & \textbf{0.76$^{\pm 0.07}$} & \textbf{1.00$^{\pm 0.00}$} & 0.77$^{\pm 0.11}$ & \textbf{-0.01$^{\pm 0.88}$} & 0.44$^{\pm 0.57}$ & -0.64$^{\pm 0.29}$ & -0.81$^{\pm 0.10}$ & 0.01$^{\pm 0.15}$ & -0.57$^{\pm 0.03}$ \\

\,\,\, + zero initialization & 0.94$^{\pm 0.17}$ & \textbf{0.94$^{\pm 0.02}$} & 0.52$^{\pm 0.14}$ & \textbf{1.00$^{\pm 0.00}$} & 0.97$^{\pm 0.03}$ & -0.74$^{\pm 0.49}$ & 0.23$^{\pm 0.27}$ & -0.54$^{\pm 0.30}$ & -0.67$^{\pm 0.06}$ & -0.13$^{\pm 0.05}$ & \textbf{-0.45$^{\pm 0.04}$} \\

\,\,\, + normalization & \textbf{1.00$^{\pm 0.00}$} & \textbf{0.94$^{\pm 0.01}$} & 0.55$^{\pm 0.09}$ & \textbf{1.00$^{\pm 0.00}$} & \textbf{0.99$^{\pm 0.01}$} & -0.03$^{\pm 0.71}$ & \textbf{0.65$^{\pm 0.30}$} & -0.32$^{\pm 0.58}$ & \textbf{-0.32$^{\pm 0.22}$} & -0.27$^{\pm 0.16}$ & -0.48$^{\pm 0.07}$ \\

\,\,\, + weighted output & \textbf{1.00$^{\pm 0.00}$} & 0.91$^{\pm 0.02}$ & 0.40$^{\pm 0.12}$ & \textbf{1.00$^{\pm 0.00}$} & 0.84$^{\pm 0.10}$ & -0.71$^{\pm 0.29}$ & 0.24$^{\pm 0.18}$ & -0.14$^{\pm 0.19}$ & -0.43$^{\pm 0.31}$ & -0.15$^{\pm 0.16}$ & -0.47$^{\pm 0.02}$ \\

\midrule
\multicolumn{5}{@{}l}{\small{\textsc{strict-small} (100M words)}}    \\[1em]

OPT\textsubscript{125M} & 0.97$^{\pm 0.01}$ & 0.58$^{\pm 0.06}$ & \textbf{0.76}$^{\pm 0.06}$ & 0.55$^{\pm 0.12}$ & \textbf{1.00}$^{\pm 0.00}$ & -0.91$^{\pm 0.10}$ & -0.98$^{\pm 0.03}$ & -0.35$^{\pm 0.17}$ & -0.73$^{\pm 0.05}$ & \textbf{-0.05}$^{\pm 0.06}$ & -0.81$^{\pm 0.08}$ \\

RoBERTa\textsubscript{base} & 0.97$^{\pm 0.02}$ & 0.49$^{\pm 0.05}$ & 0.72$^{\pm 0.12}$ & 0.93$^{\pm 0.11}$ & 0.91$^{\pm 0.08}$ & -0.99$^{\pm 0.01}$ & -0.94$^{\pm 0.04}$ & -0.30$^{\pm 0.17}$ & -0.48$^{\pm 0.08}$ & -0.37$^{\pm 0.20}$ & -0.93$^{\pm 0.10}$ \\

T5\textsubscript{base} & 0.28$^{\pm 0.04}$ & 0.25$^{\pm 0.06}$ & 0.72$^{\pm 0.03}$ & \textbf{1.00$^{\pm 0.00}$} & 0.87$^{\pm 0.03}$ & -1.00$^{\pm 0.00}$ & -0.87$^{\pm 0.05}$ & -0.39$^{\pm 0.10}$ & \textbf{-0.44}$^{\pm 0.07}$ & -0.70$^{\pm 0.10}$ & \textbf{-0.70}$^{\pm 0.05}$ \\

LTG-BERT\textsubscript{small} & \textbf{1.00$^{\pm 0.00}$} & 0.71$^{\pm 0.02}$ & 0.43$^{\pm 0.14}$ & \textbf{1.00$^{\pm 0.00}$} & 0.75$^{\pm 0.11}$ & \textbf{-0.18}$^{\pm 0.80}$ & \textbf{0.12}$^{\pm 0.21}$ & -0.48$^{\pm 0.10}$ & -0.58$^{\pm 0.04}$ & -0.48$^{\pm 0.10}$ & -0.96$^{\pm 0.04}$ \\[1em]

ELC-BERT\textsubscript{small} & \textbf{1.00$^{\pm 0.00}$} & \textbf{0.79$^{\pm 0.04}$} & 0.68$^{\pm 0.08}$ & 0.98$^{\pm 0.04}$ & 0.77$^{\pm 0.01}$ & -0.86$^{\pm 0.10}$ & 0.00$^{\pm 0.24}$ & \textbf{-0.14}$^{\pm 0.21}$ & -0.57$^{\pm 0.02}$ & -0.29$^{\pm 0.17}$ & -0.82$^{\pm 0.16}$ \\


\bottomrule
\end{tabular}
}
\caption{\label{tab:msgs}
A subset of MSGS results (defined by the Baby LM challenge) for both the models trained on 100M and 10M words. All the results indicate the model MCC or LBS for the non-control tasks. To obtain the standard deviation, each model is trained with 3 seeds and 3 learning rates for the \textsc{strict} dataset and for ELC-BERT\textsubscript{small}, the other \textsc{strict-small} datasets are trained on 5 seeds with 3 learning rates, and the average MCC/LBS is reported. The results are reported in percentage. The \textbf{bold} result indicates the best model for each dataset.
}
\end{table}

\newpage

\section{Almost Stochastic Order Significance Tests}

In this section, we put all the ASO significance tests between the backbone model LTG-BERT, ELC-BERT, and all its variations trained on the \textsc{strict} dataset for both the MSGS and GLUE benchmarks.

\subsection{GLUE - \textsc{strict} dataset}

\begin{table}[ht!]
    \resizebox{\textwidth}{!}{
        \begin{tabular}{@{}lccccc@{}}
            \toprule
            \textbf{Model} & LTG-BERT\textsubscript{base} & ELC-BERT\textsubscript{base} & zero initialization & normalized & weighted output \\
            \midrule
            LTG-BERT\textsubscript{base} & -- & 1.00 & 1.00 & 1.00 & 1.00 \\
            ELC-BERT\textsubscript{base} & 0.69 & -- & 1.00 & 1.00 & 1.00 \\
            \,\,\, + zero initialization & \textbf{0.00} & \textbf{0.05} & -- & \textbf{0.00} & \textbf{0.00} \\
            \,\,\, + normalization       & 0.90 & 1.00 & 1.00 & -- & 1.00 \\
            \,\,\, + weighted output     & 0.55 & 1.00 & 0.95 & 1.00 & -- \\
            \bottomrule
        \end{tabular}
    }
    \caption{The $\varepsilon_{\min}$ from the ASO significance test between each model on the GLUE dataset. Each row compares whether the model in the row is better than the one in the column. Results in \textbf{bold} indicate that the row model is significantly better than the one in the column.}
    \label{tab:aso_glue}
\end{table}

\subsection{MSGS - \textsc{strict} dataset}

\begin{table}[ht!]
    \resizebox{\textwidth}{!}{
        \begin{tabular}{@{}lccccc@{}}
            \toprule
            \textbf{Model} & LTG-BERT\textsubscript{base} & ELC-BERT\textsubscript{base} & zero initialization & normalized & weighted output \\
            \midrule
            LTG-BERT\textsubscript{base} & -- & 1.00 & 1.00 & 1.00 & 1.00 \\
            ELC-BERT\textsubscript{base} & \textbf{0.20} & -- & \textbf{0.28} & 1.00 & 0.83 \\
            \,\,\, + zero initialization & 0.62 & 1.00 & -- & 1.00 & 1.00 \\
            \,\,\, + normalization       & \textbf{0.01} & \textbf{0.31} & \textbf{0.02} & -- & \textbf{0.15} \\
            \,\,\, + weighted output     & \textbf{0.06} & 1.00 & \textbf{0.25} & 1.00 & -- \\
            \bottomrule
        \end{tabular}
    }
    \caption{The $\varepsilon_{\min}$ from the ASO significance test between each model on the MSGS dataset. Each row compares whether the model in the row is better than the one in the column. Results in \textbf{bold} indicate that the row model is significantly better than the one in the column.}
    \label{tab:aso_msgs}
\end{table}

\end{document}